%% file: aistats_paper.tex
\documentclass[twoside]{article}

\usepackage[accepted]{aistats2024}
\usepackage{algorithm}
\usepackage{algpseudocode}
\usepackage{algorithmicx}
\usepackage[math]{alp}
\usepackage{xcolor}
\usepackage[final]{graphicx}
\usepackage{subcaption}
\usepackage{float}
\usepackage{wrapfig}

\newcommand{\annotate}[1]{{#1}}

\newtheorem{definition}{Definition}
\newtheorem{proposition}{Proposition}
\newcommand*\formula{S}
\begin{document}

%

%

\twocolumn[

  \aistatstitle{Graph Pruning for Enumeration of Minimal Unsatisfiable Subsets}

  \aistatsauthor{ Panagiotis Lymperopoulos \And Liping Liu }
  
  \aistatsaddress{ Tufts University \And  Tufts University} ]

\begin{abstract}
  Finding Minimal Unsatisfiable Subsets (MUSes) of binary constraints is a common problem in infeasibility analysis of over-constrained systems. However, because of the exponential search space of the problem, enumerating MUSes is extremely time-consuming in real applications. In this work, we propose to prune formulas using a learned model to speed up MUS enumeration. We represent formulas as graphs and then develop a graph-based learning model to predict which part of the formula should be pruned. Importantly, our algorithm does not require data labeling by only checking the satisfiability of pruned formulas. It does not even require training data from the target application because it extrapolates to data with different distributions. In our experiments we combine our algorithm with existing MUS enumerators and validate its effectiveness in multiple benchmarks including a set of real-world problems outside our training distribution. The experiment results show that our method significantly accelerates MUS enumeration on average on these benchmark problems.
\end{abstract}

\section{Introduction}
\input{sections/introduction.tex}

\section{Related Work}
\input{sections/related_work.tex}

\section{Background}

\input{sections/background.tex}

\section{Method}
\input{sections/method.tex}

\section{Experiments}

\input{sections/experiments.tex}
\section{Conclusion}
\input{sections/conclusion.tex}

\clearpage

\bibliographystyle{plain}
\bibliography{aistats_paper}
\end{document}

%% file: sections/introduction.tex
Many problems in computer science and operations research are often formulated as constraint satisfaction problems. When a system is over-constrained and has no satisfying solutions, identifying and enumerating Minimal Unsatisfiable Subsets (MUSes) of the constraint set is one way to analyze the system and understand the unsatisfiability. Example applications include circuit error diagnosis \cite{han1999deriving}, symbolic bounded model checking \cite{clarke2000counterexample, ghassabani2017efficient} and formal equivalence checking \cite{cohen2010designers}. Additionally, MUSes may also be used to identify inconsistencies between the environment and domain knowledge \cite{goel2022rapid} in planning tasks.
A wide range of constraint satisfaction problems are represented as boolean formulas, which often take the Conjunctive Normal Form (CNF). Then, a MUS of a formula is an unsatisfiable subset of the clauses in the formula, with the added property that removing any single clause from this subset renders it satisfiable. MUS enumeration aims to find all MUSes in a formula.

However, MUS enumeration poses itself as a problem harder than a satisfiability decision problem. In practice, the enumeration problem is mainly addressed by search-based algorithms, which face a large search space containing exponentially many combinations of clauses \cite{ignatiev2015smallest}. Given the large searching space, a search often requires many steps involving calls to satisfiability solvers \cite{liffiton2016fast}. A promising direction of speeding up the enumeration is to reduce the searching space (e.g. exploring subsets that are more likely to be unsatisfiable \cite{bendik2018recursive,belov2013formula}). Likely there are still undiscovered strategies beyond manually-designed heuristics to further reduce the searching space.

Recently, neural methods have been proposed to address hard graph problems \cite{khalil2017learning, li2018combinatorial, sato2019approximation} such as the traveling salesperson problem \cite{shi2022neural} and the maximum independent set problem \cite{schuetz2022combinatorial}. Often these problems are addressed using Graph Neural Networks (GNNs) \cite{zhou2020graph}.  At the same time, there has been increasing interests in applying neural networks to problems involving logic and reasoning \cite{hitzler2022neuro}. Previous work on neural SAT-solving \cite{selsam2018learning} shows promising results in extrapolation beyond the training distribution.  We find that this parallel development presents an opportunity for MUS enumeration: we can represent a formula as a graph and then leverage the power of GNNs to accelerate MUS enumeration. As learning models, GNNs have the ability to identify patterns that are relevant to satisfiability and possibly MUSes.

In this work, we propose a learning-based pruning model, Graph Pruning for Enumeration of Minimal Unsatisfiable Subsets (GRAPE-MUST), to accelerate MUS enumeration. In particular, GRAPE-MUST represents a formula as a graph and then learns a graph neural network to prune the formula via its graph form. With this generic pruning procedure, it can be combined with most existing MUS enumeration algorithms and reduce their search spaces.

The training objective of GRAPE-MUST aims to reduce clauses in a formula while keeping it unsatisfiable. This design avoids the need of true labels in model training. We further train a GRAPE-MUST model with a large number of random formulas from our specially designed generative procedure. The trained model improves enumeration performance even in tasks without training formulas. During testing time, the pruning procedure only takes a small fraction of the overall running time that includes the enumeration time, but it speeds up the enumeration procedure significantly for a wide range of problems.

We validate the effectiveness of our approach by combining our method with existing MUS enumerators in multiple benchmarks including random formulas, graph coloring problems and problems from a logistics planning domain. We also find that GRAPE-MUST shows promising extrapolation performance on larger problems than ones it was trained on. This suggests that training GRAPE-MUST on smaller problems can be a viable strategy for accelerating MUS enumeration in larger problems. Finally, we demonstrate that GRAPE-MUST can also generalize across data distributions by improving enumeration performance in a collection of hard problems from the 2011 SAT competition MUS Track using a model trained on random formulas. This result has a strong implication: a trained GRAPE-MUST has the potential to accelerate MUS enumeration for a wide range of tasks without the need for training formulas or retraining for different tasks.

%% file: sections/related_work.tex

Finding MUSes is a well-studied problem in the field of constraint satisfaction. Even though the problem is fundamentally computationally hard \cite{ignatiev2015smallest}, its practical usefulness has motivated the development of a number of domain agnostic algorithms. These algorithms are concerned with either extracting a single MUS from a given constraint set \cite{belov2012muser2,nadel2014accelerated} or with enumerating multiple MUSes, with algorithms in the latter usually building on top of the former \cite{bendik2020must}. Since full enumeration of MUSes is often intractable, algorithms for \textit{online} enumeration \cite{liffiton2016fast,bendik2018recursive,bendik2016tunable} have been developed that promise to produce at least some MUSes in reasonable time. Our work aims to improve enumeration speed of these online algorithms in problems with boolean constraints and variables.

Recently neural methods have been applied in logical reasoning \cite{hitzler2022neuro}. Such examples include neural satisfiability solvers \cite{guo2022machine},  neural theorem provers \cite{paliwal2020graph}, and neural model counters \cite{abboud2020learning}. These methods often represent problems in graphs and apply GNNs \cite{welling2016semi} to identify structural patterns in these problems.

GNNs have been shown to be successful in addressing hard graph problems \cite{ma2021deep}. Examples include the traveling salesperson problem \cite{shi2022neural},  maximum cut and independent set \cite{schuetz2022combinatorial, toenshoff2021graph} and graph edit distance \cite{liu2022towards}. In these problems, GNNs can learn common graph patterns from a large number of problems and assist a solver in finding good solutions. They achieve this by providing search heuristics \cite{shi2022neural} or reducing the problem search space \cite{liu2022towards}.

%% file: sections/background.tex
Consider a constraint satisfaction problem in CNF over a set of boolean variables $U = \{u_i\in\{T,F\}, i =  1 \ldots N \}$ with clause set $C = \{c_i: i = 1 \ldots M\}$. Each clause $c\in C$ is a disjunction $c = \bigvee_{i=1}^n l_i$ where each literal $l_i$ is either a variable $u_i$ or the negation of a variable $\neg u_i$. The full formula $\formula = \bigwedge_{i=1}^M c_i$ is the conjunction of all clauses.

A boolean CNF formula is satisfiable if there exists an assignment to the variables in $U$ such that the formula $\formula$ evaluates to true. A boolean CNF formula is unsatisfiable if there is no such assignment to the variables.

\begin{definition}
    A Minimal Unsatisfiable Subset (MUS) of a set of clauses $C$ is a clause subset $M\subseteq C$ s.t. $M$ is unsatisfiable, and $M\backslash \{c\}$ is satisfiable $\forall c\in M$.
    \label{def:mus}
\end{definition}
A single MUS is often viewed as one minimal explanation of why $C$ is unsatisfiable. It is often desirable to enumerate multiple MUSes to locate good explanations. In this work we focus on accelerating the enumeration of MUSes, aiming specifically to aid practical applications in which full enumeration is computationally infeasible.
\begin{proposition}
    Given an unsatisfiable subset $C'\subseteq C$, if $M$ is a MUS of $C'$ then $M$ is a MUS of $C$.
    \label{prop:subset_mus}
\end{proposition}
This proposition is well known (e.g. \cite{bendik2020must}), and here we formalize it to facilitate discussion. In fact, it is commonly used in MUS enumeration algorithms that use a \textit{seed-shrink} procedure \cite{bendik2020must,liffiton2016fast,bendik2016tunable,bendik2018recursive}. This procedure looks for an initial \textit{seed} $C'\subseteq C$ that is unsatisfiable, and then repeatedly shrinks $C'$ by removing clauses until it becomes a MUS.
\begin{definition}
    A clause $c\in C'$ is critical for an unsatisfiable subset $C' \subseteq C$ if $C'\backslash \{c\}$ is satisfiable.
    \label{def:critical}
\end{definition}
Critical clauses are important in MUS enumeration as they tend to be involved in many MUSes of $C$ and some solvers use them in their search strategy \cite{bendik2018recursive}.

%% file: sections/method.tex
\label{sec:method}

\input{figures/mus_fig.tex}
In this section we develop a neural method to accelerate searching algorithms in the enumeration of MUSes. In particular, we will prune the clause set $C$ to get a smaller clause set $C'$ that is still unsatisfiable. From that point, searching algorithms can be applied to enumerate MUSes in $C'$ according to Proposition \ref{prop:subset_mus}. Our main strategy is to represent formulas as graphs and then treat the pruning problem as a node labeling problem.
\subsection{CNF formulas as attributed graphs}
\label{sec:Graph}
We represent CNF formulas as Literal Clause Graphs \cite{guo2022machine}. Consider a propositional formula $\formula=(U,C)$  in CNF with variable set $U$ and clause set $C$. To construct an attributed graph $G=(V,E,X)$ from a formula, we treat variables and their negations as one node type and clauses as another node type. We also have two types of edges: the first type connects variables to their negations, and the second type connects variables or their negations to a clause if they appear in that clause.
Formally, we construct the graph $G=(V,E,X)$ from the formula $\formula=(U,C)$ with the node set:
\begin{equation}
    \begin{split}
        V &=V_1 \cup V_2  \mbox{ with}\\
        V_1 &= \{u_i | i=1... N\} \cup \{\lnot u_i | i=1... N\},V_2 = C
    \end{split}
\end{equation}
and the edge set:
\begin{equation}
    \begin{split}
        E &= E_1 \cup E_2   \mbox{ with }\\
        E_1 &= \{(u_k, c_j) | u_k \in c_j, j=1 ... M\},\\
        E_2 &= \{(u_i,\bar{u}_i) | i=1... N\}.
    \end{split}
\end{equation}
Each node and each edge are associated with one-hot vectors indicating their types. All node and edge types are recorded respectively in two matrices  $ X = (X_V\in \{0,1\}^{|V|\times 2}, X_E \in \{0,1\}^{|E|\times 2})$ that encode the node and edge types in a one-hot manner. The graph representation $G=(V, E, X)$ keeps all information of the formula since one can recover the formula from the attributed graph. We denote the procedure by $G = \mathrm{g_{\mathrm{rep}}}(\formula)$ for easy reference later.

\subsection{CNF pruning via graph pruning}
Using our graph formulation we can achieve formula pruning through graph pruning, that is, pruning nodes in $V_2$ corresponds to removing clauses from $C$.    

We formulate the pruning problem as a node labeling problem. We learn a model $p_\theta(\by | G)$ parameterized by $\theta$, and the model predicts a vector $\by \in \{0,1\}^{M}$ of labels for nodes only in $V_2$, then $\by$ indicates \textit{which nodes to keep} after pruning. Formally, $\by$ decides the pruned subset $C' \subseteq C$.
\begin{align}
    \label{eq:prune}
    C' = \{c_i | c_i \in C, y_i = 1\}
\end{align}
From the pruned clause set $C'$, we obtain the pruned formula $\formula'=(U',C')$. Here $U'$ only contains variables involved in $C'$.

To get a differentiable training objective later, we treat the model $p_\theta(\by|G)$ as a distribution of $\by$. Here we use a simple model that treats $y_i$-s as independent Bernoulli random variables. The probabilities of $\mathbf{y}$ are computed from the input $G$ by a neural network, and $\theta$ denotes its learnable parameters. More complex $\by$ may better capture patterns in the graph but usually require more complex parameterizations and more computations.

With the pruning procedure above, the model $p_\theta(\by | G)$ essentially defines a distribution $p_\theta(\formula'|\formula)$.

\subsection{Optimization}
\label{sec:optimization}
We now need to form a training objective and learn parameters of the pruning model $p_\theta(S' | S)$. Since data labeling requires expensive searching procedures, we use a weak supervision scheme that does not need labeled data. A pruned formula $\formula'$ should be small and unsatisfiable. We first design a loss function guided by this principle.

\paragraph{Loss function:} given a formula $\formula = (U, C)$, the loss function $\mathrm{loss}(\formula'; \formula)$ computes a loss value for the pruned formula $\formula' = (U', C')$ by:
\begin{equation}
    \begin{split}
        \mathrm{loss}(\formula';\formula) & = \begin{cases}
            1                               & \text{if } \mathrm{SAT}(\formula') \\
            \left(\frac{|C'|}{|C|}\right)^2 & \text{otherwise}
        \end{cases}.
    \end{split}
    \label{eq:score}
\end{equation}
Here $\mathrm{SAT}(\formula')$ corresponds to a query to a satisfiability solver on $\formula'$ that returns true if a satisfying assignment is found and false otherwise.

The loss function equally weighs two types of undesirable pruned formulas: if $S'$ is satisfiable, then the prediction is not usable and receives a penalty of 1; and if there is no pruning and $S' = S$, again it receives a penalty of 1. Otherwise, the penalty is a function of the ratio of the number of clauses in the pruned formula to the original formula. This encourages the model to prune as many clauses as possible, while maintaining unsatisfiability. This loss function will consider critical clauses (Definition 2) automatically: removing critical clauses of $C$ produces satisfiable formulas and thus incurs high penalties. As a result, it encourages the learning model to shrink the formulas while keeping these clauses intact. While this loss does not penalize destroying MUSes, it avoids searching for MUSes during training and thus enables better scalability. We further discuss this issue in later sections.

\paragraph{Learning objective:} the loss function $\mathrm{loss(\formula'; \formula)}$ is not differentiable with respect to $\formula'$, and there is not a straightforward continuous relaxation of it. To get a differentiable learning objective, we take the expectation of the loss using the distribution $p_\theta(\formula'| \formula)$.
\begin{equation}
    L(\theta;\formula) = \mathbb{E}_{\formula'\sim p_\theta(\formula'|\formula)}[\mathrm{loss}(\formula'; \formula)].
    \label{eq:objective}
\end{equation}
Then we can draw a few Monte Carlo samples of $S'$ and estimate the gradient with respect to $\theta$ through the score function estimator \cite{williams1992simple}.
\begin{align*}
    \nabla_\theta L(\theta;\formula) & = \mathbb{E}_{\formula'\sim p_\theta(\formula'|\formula)}[\mathrm{loss}(\formula'; \formula) \cdot \nabla_\theta \log p_\theta(\formula'|\formula) ].
\end{align*}
Though the score function estimator often has large variance, it works well in our experiments. We will explore techniques to reduce the variance in the future.

\subsection{Model architecture}
We now present our implementation of the graph pruning model $p_\theta(\by | G)$ with GNNs. To facilitate the scalability of our method, we use a lightweight architecture with a relatively small memory footprint. Along with node features $X_v$ indicating node types, we also append random node features to improve the expressiveness of the network \cite{abboud2021surprising,sato2021random}. So the input to the GNN is:
\begin{align}
    H^0 = [X_V, R], ~~~ R \in \bbR^{N \times d_r} \sim \mathcal{N}(\bzero, \bI).
\end{align}
Then we use an $L$-layer GNN with heterogeneous message passing layers \cite{wang2022survey} to compute node representations $Z \in \bbR^{N\times d_o}$ for each node type.
\begin{align}
    Z = \mathrm{gnn}(G, H_0).
\end{align}
Here $\mathrm{gnn}(\cdot)$ denotes the function of the GNN. Finally, we use a simple MLP to predict probabilities of node labels $\by$, which indicate which clauses in $C$ to \textit{keep} in the pruned formula.
\begin{align}
    \bmu & = \mathrm{mlp}(Z), ~~~~
    \by   \sim \mathrm{Bernoulli}(\bmu).
\end{align}
Here the MLP applies to each node representations to compute a probability value.

We take a ``conservative-to-aggressive'' strategy to train the model. We initialize the bias in the last layer of the MLP to a moderate negative value (e.g.  $-3$) and network weights to small values. This results in the model initially assigning a pruning probability around $0.05$ to all nodes, which means the initial model does little pruning to all formulas. Then as the model learns to minimize the loss, it becomes more aggressive and prunes formulas to get smaller unsatisfiable formulas. This strategy avoids initial models that are too aggressive and cannot get unsatisfiable formulas as such models cannot improve through small updates and are hard to optimize.

\subsection{Randomized formula generation from problem statistics}
\label{sec:random_formula_gen}
For tasks without training formulas, we generate random formulas as the training data to train our model. However, it is non-trivial to generate formulas that are like real problems. Pure random formulas that are unsatisfiable tend to have small MUSes.

In this work, we devise a randomized procedure that generates formulas with similar clause lengths and clause-to-variable ratio with that from target tasks. We also consciously try to control sizes of MUSes in these formulas. According to the target clause-to-variable ratio, we first decide the number of variables and a lower bound of the number of clauses. The generation procedure then proceeds by sampling one clause at a time and adding it to the formula only if the resulting clause-set is satisfiable.
\input{algorithms/bin_search_prune.tex}
This procedure is repeated until the lower bound is reached. After that point, we continue to add clauses until the formula becomes unsatisfiable. The literals in each clause are uniformly sampled, and the length of the clause is randomly decided according to the target clause length distribution.

This procedure yields problems that resemble the data distribution in clause lengths and clause-to-variable ratios. Without satisfiability checking, random clauses tend to make MUSes smaller. Our procedure guarantees satisfiability initially and thus tends to generate formulas with larger MUSes than pure random formulas with same lengths. The procedure does require a large number of calls to a SAT solver, but many problems can be generated in parallel and we only need to generate a training set once for a wide range of problems.

\subsection{Test-time pruning}

In testing time, we use a deterministic procedure to compute a valid pruning vector $\by$ to avoid randomness.  We apply a threshold $t$ to truncate the probability vector $\bmu$ to get $\by$. Then we check whether the pruned formula $\formula'$ from $\by$ is satisfiable or not. We then proceed to search for the smallest threshold value (most aggressive pruning) that yields an unsatisfiable formula using binary search. The search is conducted over threshold values from $\frac{\max(\bmu)}{k}$ to $\max(\bmu)$ with $k$ being a hyperparameter not related to the size of $\formula$.  This procedure is formally described by Algorithm \ref{alg:greedy_prune_binsearch}.  In the worst case, $t=\max(\bmu)$ will give $\formula'=\formula$ without pruning. There are $O(\log k)$ SAT calls, which is typically much less than SAT calls in MUS searching algorithms.

After we have obtained the pruned formula $\formula'$, we run a MUS enumeration algorithm on $\formula'$ to enumerate MUSes of $\formula$. The main gain is that time saved by running the enumeration algorithm on a smaller formula $\formula'$. While some MUSes may be destroyed during pruning, we deem this a reasonable compromise as in practical problems enumerating all MUSes is already prohibitively expensive. Our experiments show that our pruning allows for more MUSes to be found within the same time-limit in both synthetic and real-life problems.

%% file: figures/mus_fig.tex
\begin{figure*}
    \centering
    \includegraphics[width=0.7\textwidth]{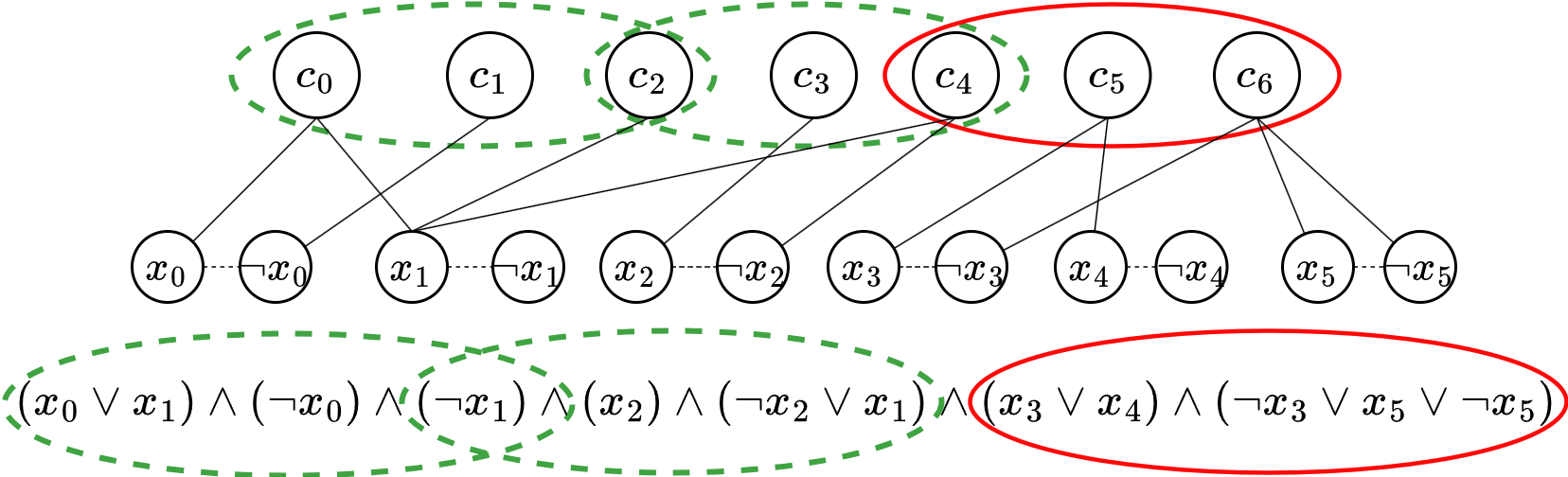}
    \label{fig:fprune}
    \caption{Graph pruning for MUS enumeration. CNF formulas are represented as Literal Clause Graphs and the clause nodes are pruned. The resulting graph is converted back into a formula and MUSes are enumerated. Green denotes two MUSes in the formula. Red denotes clauses that can be pruned without affecting the MUSes. $c_2$ is a critical constraint}
\end{figure*}

%% file: algorithms/bin_search_prune.tex
\begin{algorithm}[H]
    \footnotesize
    \caption{Prune a CNF formula}\label{alg:greedy_prune_binsearch}
    \textbf{Input} $\formula=(U,C)$, $\bmu$, k\\
    \textbf{Output} $\formula'=(U',C')$\\
    \begin{algorithmic}[1]
        \State $t_{max} \gets \max(\bmu)$
        \State $t_{min} \gets \min(\bmu)$
        \While {$t_{max}-t_{min} > 0$}
        \State $t \gets (t_{max}-t_{min})/2$
        \State $\by \gets \bmu \le t$ 
        \State $\formula' \gets Prune(\formula,\by)$ \Comment{using eq. \ref{eq:prune}}
        \If{$SAT(\formula')$}
        \State $t_{min} \gets t$
        \Else
        \State $t_{max} \gets t$
        \EndIf
        \EndWhile\\
        \Return $\formula'$
    \end{algorithmic}
\end{algorithm}

%% file: sections/experiments.tex
In this section, we evaluate the effectiveness of the proposed pruning strategy in MUS enumeration tasks. We first check whether applying a pruning model before an enumeration algorithm improves the algorithm's performance in enumerating MUSes. We also investigate the generalization ability of our model by evaluating a trained pruning model on formulas from a distribution different from the training distribution. Finally, we evaluate the feasibility of using a model trained on random formulas on a benchmark of challenging MUS enumeration problems from the literature, thus reducing the need to train a model on each new problem distribution.
\paragraph{Datasets:} we evaluate GRAPE-MUST on four datasets including both randomly generated and real-world problems. We use the following datasets:

\textit{Random Formulas.} We generate formulas using the procedure described in \cite{selsam2018learning}, resulting in formulas with about 700 clauses. Exact generation parameters are \annotate{available in the appendix}.

\textit{Logistics Planning.} We use a standard logistics planning problem with variable numbers of cities, addresses, airplanes, airports and trucks. We create random initial and goal states and also vary the number of deliveries in a given timeframe. We only keep infeasible problems and then use MADAGASCAR \cite{rintanen2014madagascar} to translate our planning problems into boolean CNF formulas. The derived formulas have about 800-1200 variables and 8000-15000 clauses. Exact generation parameters, domain file and conversion parameters \annotate{are available in the appendix}.
\input{tables/random_formulas.tex}
\input{tables/logistics.tex}

\textit{Graph Coloring.} To generate random graph coloring problems, we first sample a random graph with 10 to 30 nodes from the Erd\H{o}s-R\'{e}nyi model with edge probability 0.8,  and then randomly choose a color number between 4 and 7. Then, we convert formulas to SAT using a standard translation procedure \annotate{described in the appendix}. We only sample unsatisfiable formulas by discarding any satisfiable ones. The resulting formulas have up to 210 variables and up to about 2500 clauses.

\textit{Hard problems from SAT Competition 2011 MUS track.}\footnote{http://www.cril.univ-artois.fr/SAT11/}. This benchmark contains problems from various applications such as planning, software and hardware verification that vary in size from a few hundred to millions of clauses and variables. We limit our investigation to problems in the benchmark that we deem hard: They contain at least $10^5$ clauses and a state-of-the-art enumerator identifies at most $10^3$ MUSes within a 2 hour time limit without exhausting the number of MUSes in the formulas. Given these criteria, we evaluate GRAPE-MUST on 63 problems from this benchmark.

For each of the first three datasets, we test enumeration algorithms on 500 randomly generated problems and repeat each experiment 5 times. As a separate note, these problems do not pose significant difficulties to modern SAT solvers such as Glucose-3.0 \cite{audemard2009glucose}, which can evaluate their satisfiability within 10 milliseconds.

\textbf{Enumeration algorithms:} we apply the pruning strategy to three contemporary online MUS enumeration algorithms: \textit{MARCO} \cite{liffiton2016fast}, \textit{TOME} \cite{bendik2016tunable}, and \textit{REMUS} \cite{bendik2018recursive}. All three algorithms are available as part of the MUST \cite{bendik2020must} toolbox.

\textbf{Model hyperparameters:} we use the GraphConv operator \cite{morris2019weisfeiler} to implement message passing in heterogeneous graphs formed by formulas. In all experiments we use five message passing layers with 64 hidden units. We use two layers for the MLP with ReLU activations. We train the models for a maximum of 2 million formulas with early stopping and use the Adam optimizer with a learning rate of 0.0001 and a batch size of 32. At test time, we set $k=10$ in algorithm \ref{alg:greedy_prune_binsearch} to compute the pruning. All our experiments are carried out on a server with 4 NVIDIA RTX 2080Ti GPUs, and an Intel(R) Core(TM) i9-9940X processor with 130 GB of memory.


\subsection[short]{Improving enumeration performance}
\label{sec:performance_comparison}
In this section we present the results of our first experiment. We run three existing MUS enumeration algorithms and compare performances of each algorithm with and without our pruning step. We run all three algorithms with the same default parameters in all problems. We evaluate the algorithms on the three aforementioned datasets. The evaluation metric is the average number of MUSes enumerated within a fixed time budget: the larger, the better. The running time of GRAPE-MUST on GPUs is included within the time-budget. In our experiments, we record three average numbers for three time budgets: 1, 2, and 5 seconds. We also run smaller scale experiments for a 30 minute time-budget and present the results as well as pruning statistics in the appendix. We note that these classes of problems are not very challenging to MUS enumerators at the problem size and time limit used in this experiment. Still, we believe that the results are indicative of the effectiveness of pruning in accelerating MUS enumeration. Performance improvement in more challenging problems is shown in following sections.

\input{tables/graph_color.tex}
\input{figures/generalization.tex}

The experiment results from the three datasets are tabulated in Table \ref{tab:random_formulas},  \ref{tab:logistics}, and \ref{tab:graph_color}. These results show that GRAPE-MUST allows MARCO and REMUS to find more MUSes on all three datasets. On the \textit{Graph Coloring} dataset, REMUS with pruning nearly doubles the number of MUSes found by REMUS alone across all timeouts. It also helps TOME to find more MUSes on two datasets, \textit{Random Formulas} and \textit{Logistics Planning}. Only on the \textit{Graph Coloring} dataset, pruning actually harms the performance of TOME. Importantly REMUS is the strongest algorithm among the three \cite{bendik2018evaluation}, and our pruning model improves REMUS on all three datasets, making GRAPE-MUST+REMUS the strongest configuration.

Further analysis shows how the three algorithms benefit differently from a prior pruning step.
REMUS invests significant computation in finding small unsatisfiable subsets and heavily depends on critical constraints. GRAPE-MUST learns to prune non-critical constraints, which makes finding and using critical constraints easier for REMUS. Compared to REMUS, MARCO benefits less from pruning in all three tasks. MARCO's search strategy tends to start from large unsatisfiable subsets and shrinks them to find MUSes with many SAT calls. Since single SAT calls are in practice not as expensive in boolean problems, MARCO's less aggressive seed-searching strategy means it does not benefit as much from pruning as REMUS. TOME builds chains of subsets of the formula and looks for the smallest unsatisfiable one in the chain \cite{bendik2016tunable}. The sparsity induced by pruning may make it harder for TOME to find fully unsatisfiable chains, requiring it to perform binary search or build new chains more often. It is also important to note that TOME enjoys the weakest negative correlation between number of constraints and MUSes enumerated \cite{bendik2018evaluation}, making positive effects of pruning easier to mask.

These results, consistent with the 30-minute runs shown in the appendix, indicate that while GRAPE-MUST is beneficial to MUS enumeration in many cases, the choice of underlying solver matters and REMUS is consistently the best choice in our problems.

\subsection[short]{Extrapolation to larger problems}
\label{sec:generalization}
In this experiment we investigate the extrapolation capability of our model to see if it can successfully prune formulas of larger sizes than the ones used for training.
\paragraph{Experiment Settings:} we use the same model trained on randomly generated formulas with 100 variables as in section \ref{sec:performance_comparison}. We evaluate the model on formulas with 100, 150, 200, 250 and 300 variables. We evaluate the solvers on 500 formulas of each size and measure the absolute and relative improvement in MUS enumeration performance.

\paragraph{Results:} Figure \ref{fig:generalization} shows the relative and absolute improvement in MUS enumeration for each solver using GRAPE-MUST. Pruning generalizes well to larger problems, with the average problem size reduction only decreasing by about $7$ percentage points from 100 to 300 variables. A table with size reduction figures and enumeration results is available in the appendix.

Consistent with previous results, pruning benefits REMUS significantly more than the other solvers, and here, MARCO benefits the least. Particularly, figure \ref{fig:generalization} (left) indicates that for REMUS the improvement is largest in formulas of around 200 clauses. However, in very large formulas the effect of pruning diminishes as even pruned problems become too large for the 5-second timeout. As shown in \ref{fig:generalization} (right), the number of MUSes found by all solvers in the largest formulas within the time limit is very small, resulting in large variance in the effect of pruning. Nevertheless, the results up to 200-250 formulas suggest that training GRAPE-MUST on smaller problems can be a viable strategy for improving the MUS enumeration performance of REMUS even in larger instances.

\subsection[short]{Performance improvement in benchmark problems}
\label{sec:pretrained}
In this experiment we evaluate a model trained on randomly generated formulas on a collection of hard problems from the 2011 SAT competition MUS enumeration benchmark problems.

\paragraph{Experiment Settings:} we scale up our model to use 6 hidden layers and a latent dimension of 128 units. We train the model on random formulas generated as described in section \ref{sec:random_formula_gen}. We obtain the dataset statistics from the entire SAT Competition 2011 MUS track problem set and no additional information from the dataset is used for training. We train the model on 2 million formulas with 50 to 10000 variables and use $k=100$ to compute the pruning at test time.  We evaluate the model on 63 problems as described in the beginning of this section. We compare the performance of the highest performing enumerator \cite{bendik2020must} REMUS with and without pruning using a timeout of 2 hours.
\input{figures/hard_problems.tex}

\paragraph{Results:}
Figure \ref{fig:hard_problems} summarizes the performance of the two methods in the benchmark. Out of 63 problems, GRAPE-MUST enables the discovery of more MUSes in 30 problems, results in no change in 21 and in performance decrease in 12 problems. Interestingly in 2 problems in which REMUS alone finds no MUSes within the time limit, GRAPE-MUST enables the enumeration of 2349 and 30 MUSes respectively.

On average, GRAPE-MUST removes 9444 clauses from the problems, which corresponds to about $1\%$ of the problem size. Compared to previous experiments this is a small reduction (see appendix). However, pruning $1\%$ of randomly chosen clauses from the benchmark problems invariably yields satisfiable problems. This suggests that GRAPE-MUST is able to identify non-critical constraints in the benchmark problems despite the difference from the training distribution. In almost all failure cases, the pruning removes less than $0.1\%$ of the clauses, indicating that the effort put into pruning the problems had minimal effect, only taking time away from MUS enumeration. Furthermore, comparisons against naive pruning heuristics (see appendix), show that GRAPE-MUST is more general and robust in real-world problems.

Overall this experiment indicates that training GRAPE-MUST on random formulas is a viable strategy for accelerating MUS enumeration in difficult real-world problems. We are therefore releasing the trained GRAPE-MUST model along with the training code.

%% file: tables/random_formulas.tex
\begin{table*}[t]
    \footnotesize
    \centering
    \scalebox{0.9}{
        \begin{tabular}{|c|c|c|c|}
            \hline
            \textbf{Solver}    & 1 (s)                      & 2 (s)                       & 5 (s)                       \\
            \hline
            MARCO              & 230.85$\pm$ 8.79           & 465.24$\pm$ 16.41           & 1145.92$\pm$ 36.0           \\
            GRAPE-MUST + MARCO & \textbf{231.35$\pm$ 8.32}  & \textbf{469.24$\pm$ 15.5}   & \textbf{1157.69$\pm$ 34.03} \\
            \hline
            REMUS              & 891.28$\pm$ 29.83          & 1933.27$\pm$ 60.64          & 5188.03$\pm$ 149.6          \\
            GRAPE-MUST + REMUS & \textbf{1171.74$\pm$ 33.1} & \textbf{2400.03$\pm$ 64.73} & \textbf{6055.5$\pm$ 152.93} \\
            \hline
            TOME               & 156.08$\pm$ 5.25           & 303.58$\pm$ 9.89            & 723.29$\pm$ 23.32           \\
            GRAPE-MUST + TOME  & \textbf{175.9$\pm$ 6.0}    & \textbf{347.56$\pm$ 11.74}  & \textbf{848.8$\pm$ 27.96}   \\
            \hline
        \end{tabular}
    }
    \caption{Average number of MUSes enumerated for random problems in 1, 2 and 5 seconds for different solvers with and without pruning. Bold indicates higher average. }
    \label{tab:random_formulas}
\end{table*}

%% file: tables/logistics.tex
\begin{table*}[t]
    \footnotesize
    \centering
    \scalebox{0.9}{
        \begin{tabular}{|c|c|c|c|}
            \hline
            \textbf{Solver}    & 1 (s)                      & 2 (s)                      & 5 (s)                       \\
            \hline
            MARCO              & 148.45$\pm$ 16.72          & 288.07$\pm$ 31.17          & 648.67$\pm$ 65.36           \\
            GRAPE-MUST + MARCO & \textbf{163.58$\pm$ 20.34} & \textbf{312.88$\pm$ 37.33} & \textbf{680.44$\pm$ 75.44}  \\
            \hline
            REMUS              & 151.42$\pm$ 14.19          & 322.12$\pm$ 31.43          & 814.14$\pm$ 77.16           \\
            GRAPE-MUST + REMUS & \textbf{245.25$\pm$ 32.96} & \textbf{535.82$\pm$ 72.2}  & \textbf{1321.94$\pm$ 160.1} \\
            \hline
            TOME               & 56.25$\pm$ 3.68            & 109.92$\pm$ 6.76           & 253.19$\pm$ 13.64           \\
            GRAPE-MUST + TOME  & \textbf{67.85$\pm$ 5.73}   & \textbf{130.8$\pm$ 10.77}  & \textbf{313.76$\pm$ 24.07}  \\
            \hline
        \end{tabular}
    }
    \caption{Average number of MUSes enumerated for logistics planning problems in 1, 2 and 5 seconds for different solvers with and without pruning. Bold indicates higher average.}
    \label{tab:logistics}
\end{table*}

%% file: tables/graph_color.tex
\begin{table*}[t]
    \footnotesize
    \centering
    \scalebox{0.9}{
        \begin{tabular}{|c|c|c|c|}
            \hline
            \textbf{Solver}    & 1 (s)                      & 2 (s)                       & 5 (s)                        \\
            \hline
            MARCO              & 113.97$\pm$ 13.59          & 196.27$\pm$ 24.27           & 439.55$\pm$ 56.51            \\
            GRAPE-MUST + MARCO & \textbf{119.62$\pm$ 16.92} & \textbf{231.34$\pm$ 33.39}  & \textbf{514.68$\pm$ 74.22}   \\
            \hline
            REMUS              & 216.13$\pm$ 24.05          & 423.62$\pm$ 50.3            & 971.27$\pm$ 111.76           \\
            GRAPE-MUST + REMUS & \textbf{428.55$\pm$ 65.89} & \textbf{958.15$\pm$ 145.02} & \textbf{2371.07$\pm$ 358.77} \\
            \hline
            TOME               & \textbf{115.96$\pm$ 13.99} & \textbf{195.96$\pm$ 22.96}  & \textbf{385.97$\pm$ 42.59}   \\
            GRAPE-MUST + TOME  & 81.39$\pm$ 11.35           & 154.72$\pm$ 21.89           & 339.01$\pm$ 46.81            \\
            \hline
        \end{tabular}
    }
    \caption{Average number of MUSes enumerated for graph coloring problems in 1, 2 and 5 seconds for different solvers with and without pruning. Bold indicates higher average.}
    \label{tab:graph_color}
\end{table*}

%% file: figures/generalization.tex
\begin{figure*}[h]

  \begin{subfigure}{0.5\textwidth}
    \centering
    \includegraphics[width=0.925\linewidth]{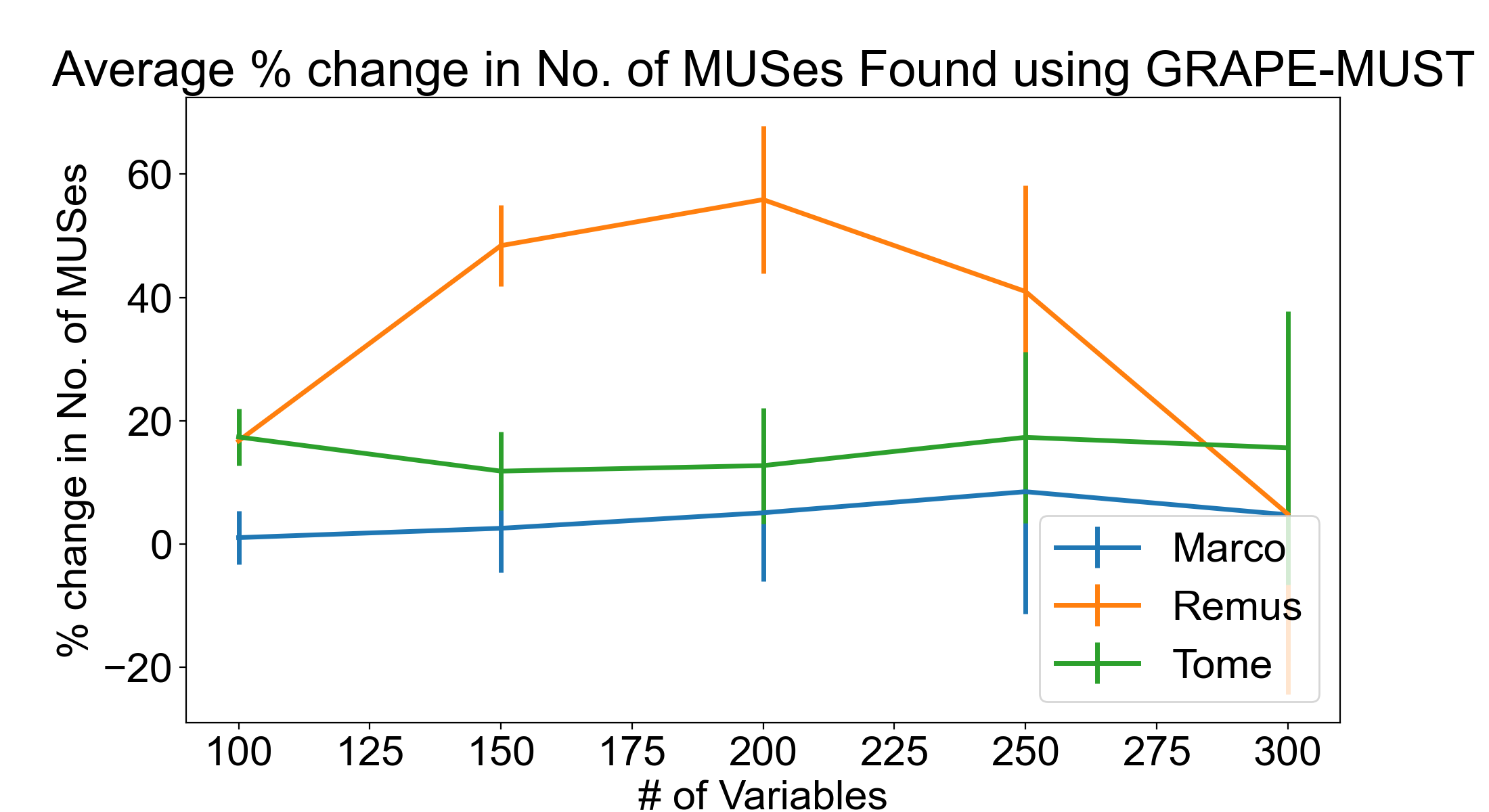}
    \caption{Percent improvement in MUS enumeration}
    \label{fig:perc_change}
  \end{subfigure}%
  \begin{subfigure}{0.5\textwidth}
    \centering
    \includegraphics[width=0.925\linewidth]{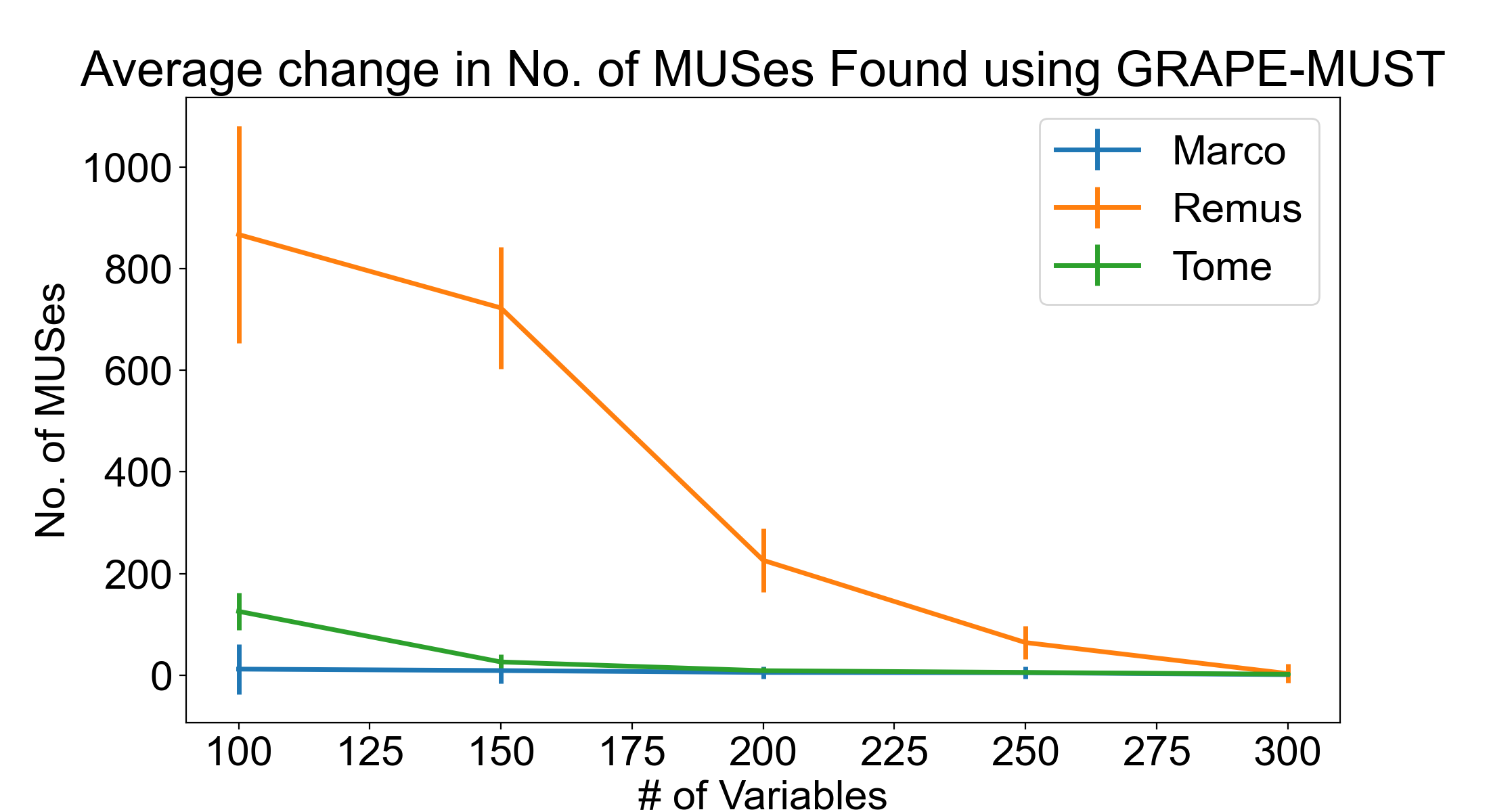}
    \caption{Absolute improvement in MUS enumeration}
    \label{fig:abs_change}
  \end{subfigure}
  \caption{Extrapolation to larger formulas. Relative (a) and absolute (b) improvement over the base solvers using GRAPE-MUST on formulas of increasing size at a 5-second timeout. Despite the distribution shift, REMUS benefits proportionally more from pruning in formulas up to 200 variables as the smaller size of the pruned formulas accelerates enumeration. MARCO and REMUS maintain a small improvement across all formulas. As indicated by (b) the variance in (a) increases rapidly with formula size as the actual number of MUSes found becomes very small.}
  \label{fig:generalization}
\end{figure*}

%% file: figures/hard_problems.tex
\begin{figure}[t]
    \centering
    \includegraphics[width=0.35\textwidth]{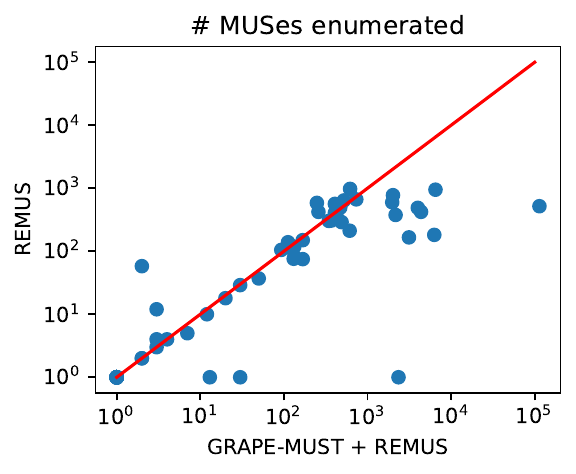}
    \caption{Comparison between REMUS and GRAPE-MUST+REMUS on hard benchmark problems. Red line indicates equal number of MUSes. Overall GRAPE-MUST enables the discovery of more MUSes.}
    \label{fig:hard_problems}
\end{figure}

%% file: sections/conclusion.tex
In this work we have introduced a method that uses learning-based graph pruning to accelerate the enumeration of MUSes from unsatisfiable CNF formulas. The main approach is converting CNF formulas to graphs and then formulating the pruning problem as a node labeling problem. We have also designed a loss function and a differentiable objective to train a pruning model. Extensive experimental results show that MUS enumeration algorithms benefit from pruning in most cases, despite the possibility of destroying some MUSes. The learned model is also able to extrapolate to problems larger than ones in its training data. Finally, GRAPE-MUST trained on random formulas is able to generalize across data distributions, improving MUS enumeration performance on hard real-world problems without the need of large datasets.